\documentclass[10pt,twocolumn,letterpaper]{article}

\usepackage{3dv}
\usepackage{times}
\usepackage{epsfig}
\usepackage{graphicx}
\usepackage{amsmath}
\usepackage{amssymb}
\usepackage{url}
\newcommand*\samethanks[1][\value{footnote}]{\footnotemark[#1]}

\usepackage{times}
\usepackage{url}
\usepackage{graphicx}
\usepackage{booktabs}
\usepackage{algorithm}
\usepackage{algorithmic}
\usepackage{adjustbox}
\usepackage{enumitem}
\usepackage{amsmath}
\usepackage{amsfonts}
\usepackage{multirow}
\usepackage{pifont}
\usepackage{dblfloatfix}

\newcommand{\vect}[1]{\mathbf{#1}}

\def\eg{\textit{e.g.}~}
\def\ie{\textit{i.e.}}

\def\etal{\textit{et al.}}

\usepackage{tabularx}

\newcommand{\tabincell}[2]{\begin{tabular}{@{}#1@{}}#2\end{tabular}}

\threedvfinalcopy 

\def\threedvPaperID{212} 

\ifthreedvfinal\fi
\begin{document}
\title{Structured Coupled Generative Adversarial Networks for \\ Unsupervised Monocular Depth Estimation}

\author{
    2D and 3D Computer Vision
    \and
    Perception \\
}

\author{Mihai Marian Puscas$^{1\dagger}$\thanks{Equal contribution. \ $^\dagger$Work performed while at SAP ML Research Berlin} \ \ \
        ~Dan Xu$^3$\samethanks \ \ \
        ~Andrea Pilzer$^2$\samethanks \ \ \
        ~Niculae Sebe$^{1, 2}$ \vspace{5pt}\\
    $^1$Huawei Technologies Ireland, Dublin, Ireland\\
    $^2$DISI, University of Trento, Povo (TN), Italy\\
    $^3$Department of Engineering Science, University of Oxford, Oxford, UK \\
{\tt\small mihai.puscas@huawei.com, \{andrea.pilzer, niculae.sebe\}@unitn.it, danxu@robots.ox.ac.uk}
}

\maketitle

\begin{abstract}
Inspired by the success of adversarial learning, we propose a new end-to-end unsupervised deep learning framework for monocular depth estimation consisting of two Generative Adversarial Networks (GAN), deeply coupled with a structured Conditional Random Field (CRF) model. The two GANs aim at generating distinct and complementary disparity maps  
and at improving the generation quality via exploiting the adversarial learning strategy. The deep CRF coupling model is proposed to fuse the generative and discriminative outputs from the dual GAN nets. As such, the model implicitly constructs mutual constraints on the two network branches and between the generator and discriminator. This facilitates the optimization of the whole network for better disparity generation. Extensive experiments on the KITTI, Cityscapes, and Make3D datasets clearly demonstrate the effectiveness of the proposed approach and show superior performance compared to state of the art methods. The code and models are available at~\url{https://github.com/mihaipuscas/3dv---coupled-crf-disparity}.
\end{abstract}

\vspace{-7pt}
\section{Introduction}
\vspace{-0pt}
Estimating scene depth from monocular images is a fundamental task in computer vision which can be potentially applied in various applications such as autonomous driving~\cite{behl2017bounding}, Visual SLAM~\cite{turan2017non}. 
The main drawback of supervised-based systems is their dependence on costly depth-map annotations.
As such, researchers have proposed unsupervised-based deep models using self-supervised view synthesis based on photometric error estimation~\cite{garg2016unsupervised,godard2016unsupervised}. 
Within this pipeline, the quality of the view synthesis directly affects the performance of the final depth prediction. Adversarial learning has been introduced to improve the synthesis process in depth estimation systems~\cite{adadepth,pilzer2018unsupervised} by simply adding a frame-level discriminative loss for the image synthesis. However, the depth prediction maps and the discriminative error maps share meaningful structural information, e.g., objects in the input images are recognizable in both maps, and similar/close regions with higher generative errors tend to output higher discriminative errors. These structured relationships cannot be directly modeled in a standard GAN as the generator and the discriminator are not directly connected and thus do not explicitly flow gradients between them during the network optimization process. We argue that the discriminative and generative sub-networks hold complimentary structural information and jointly modeling it 
leads to a concurrent refinement of the produced discriminative error maps and the disparity maps used in the synthesis process, further improving the learned depth prediction model.




\begin{figure}[!t]
\centering
\includegraphics[width=0.99\linewidth]{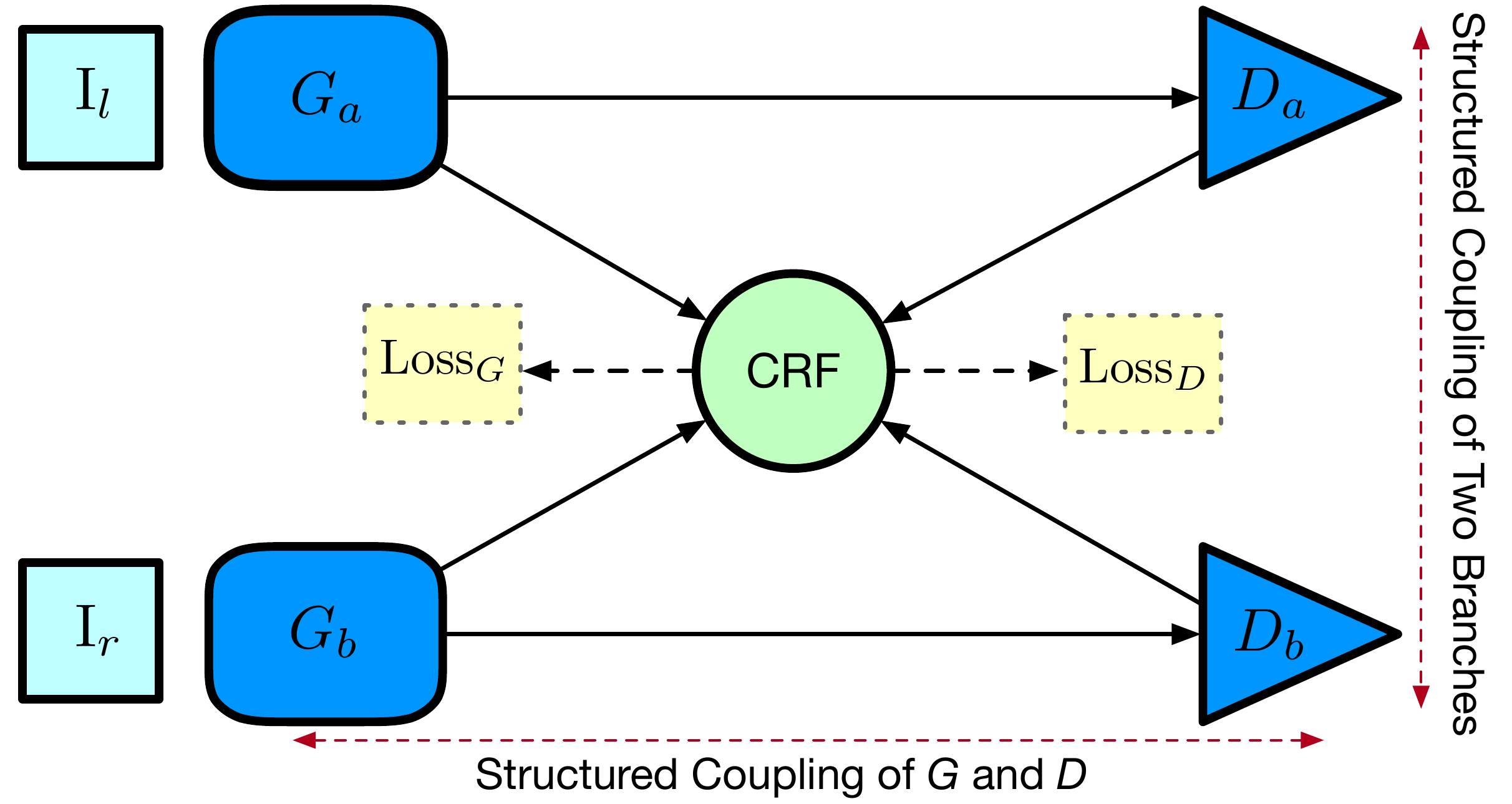} 
\vspace*{-0mm}
\caption{Illustration of the proposed structured coupling approach for adversarial monocular depth estimation. $G_a, G_b$ and $D_a, D_b$ denote generators and discriminators respectively.}
\label{motivation}
\vspace{-10pt}
\end{figure}

In this paper, we propose a structured adversarial deep model for unsupervised monocular depth estimation. The model consists of a dual generative adversarial network (DGAN), which takes stereo images as input and performs image synthesis with the two branches containing separate generators and discriminators formulated as GANs~\cite{goodfellow2014generative}. The produced disparity maps are used to synthesize images from a single view. The complimentary stereo information is learned by a hallucinatory sub-network such that during inference the system can operate in a monocular fashion. 

We further propose a deep CRF model to couple the network on two levels: We bind the two branches corresponding to each stereo image together, such that the complimentary stereo information is modeled. At the same time we model the complimentary structured information observed between the synthesized depth maps and discriminative error maps, through the linkage of the generative and discriminative sub-networks (see Fig.~\ref{motivation}). This 2D linkage constrains the generative process through the use of a structured error, allowing for a structured refinement of the final synthesized depth map. The learning of the CRF model is thus jointly determined by the errors from both the generators and the discriminators.

We show how the proposed coupled CRF model can be solved with the mean-field theory, and present a neural network implementation for the inference, enabling the joint optimization with the backbone DGAN in an end-to-end fashion. In the testing phase, only one single image is required. To summarize, our main contribution is threefold: 
\begin{itemize}[leftmargin=*]
\item We propose a novel CRF coupled Dual Generative Adversarial Network (CRF-DGAN) for unsupervised monocular depth estimation, which implicitly explores making the adversarial and structured learning benefit each other in an unified deep model for the task.  

\item Our model contains a dual GAN structure able to exploit the inherent relationship between stereo images to better learn the disparity maps. A coupled CRF model, implemented as a CNN, is presented to provide a structured fusion of the two sub-networks, as well as a structured connection between the discriminator and the generator. 

\item We conduct extensive experiments on the KITTI, Cityscapes, and Make3D datasets, clearly demonstrating the advantage of structured coupling in the designed dual GAN networks for the monocular depth estimation task. The proposed model is potentially useful for other GAN based applications possessing rich structural information. A very competitive performance is reported on KITTI as compared to state-of-the-art methods. The code will be made publicly available upon acceptance.
\end{itemize}

\section{Related Work}
\vspace{-5pt}
We review the related works in monocular depth estimation from four research directions: supervised based methods, unsupervised based methods, probabilistic graphical models based methods, and GAN based methods.

\par\noindent\textbf{Supervised Learning Methods.}
CNN based supervised monocular depth estimation systems, requiring single view images and corresponding ground-truth depth maps, have been widely developed~\cite{eigen2014depth,laina2016deeper}. Among them,~\cite{eigen2014depth} propose a coarse-to-fine network structure using multi-scale deep features. \cite{laina2016deeper} further introduce a deeper network designed for the task able to recover good scene details. Due to the high cost in collecting depth maps, some methods~\cite{adadepth} consider using additional synthetic data to facilitate the model optimization. 

\begin{figure*}[!t]
\centering
\includegraphics[width=1\linewidth]{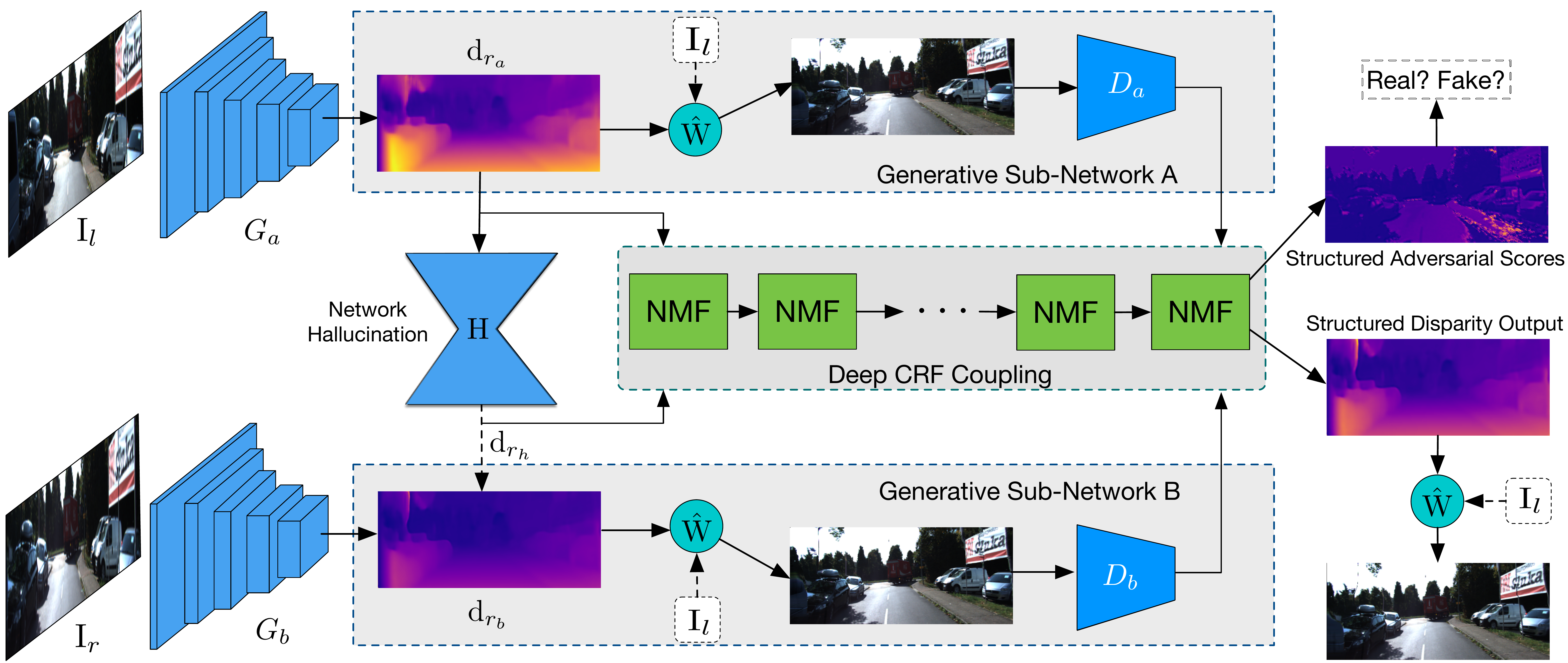} 
\vspace*{-8pt}
\caption{Framework overview of the proposed CRF-DGAN for unsupervised monocular depth estimation. $\mathrm{\hat{W}}$ is a warping operation to obtain a synthesized image. $D_a$ and $D_b$ are two discriminators corresponding to the two generative sub-networks. NMF denotes the neural network implementation of the continuous mean-field updating which composes the deep CRF model for structured coupling of the dual GANs. The training phase utilizes a pair of stereo images $\vect{I}_l$ and $\vect{I}_r$ as input, while in the testing phase, only one single image is required.}
\label{framework}
\vspace{-8pt}
\end{figure*}

\par\noindent\textbf{Unsupervised Learning Methods.}
Unsupervised depth estimation methods generally learn the depth estimation in a self-supervised fashion with a view reconstruction loss, as in  ~\cite{mahjourian2018unsupervised}. Recently, several research works have been proposed in this direction~\cite{kuznietsov2017semi,wang2017learning,zhan2018unsupervised,pilzer2018unsupervised,guo2018learning}. \cite{garg2016unsupervised} conduct a pioneering exploration and propose a deep learning framework considering multiple geometric constrained losses. Following~\cite{garg2016unsupervised},~\cite{godard2016unsupervised} propose a left-right consistency network design to synthesize images from both views to obtain stronger supervision. 
There exist some other works~\cite{wang2017learning} which jointly learn depth estimation and ego-motion, or to refine the predicted depth~\cite{pilzer2019cvpr}. However, none of these works consider utilizing the adversarial learning strategy for better view synthesis thus improving the depth estimation.

\par\noindent\textbf{Probabilistic Graphical Models.}
Conditional Random Fields have been utilized in several methods~\cite{liu2016learningtpami,xu2017multi,xu2018monocular}, performing a structured refinement of the depth estimation maps. The CRF's use is built upon the idea that pixels corresponding to close appearance or spatial regions will likely have similar depths.~\cite{liu2016learningtpami} propose a continuous CRF model for the task.~\cite{xu2017multi} propose a multi-scale fusion guided by CRF to improve depth estimations. In this work, we consider a different way via constructing a CRF model to perform structured coupling of a dual GAN network to benefit from the advantages of both.

\par\noindent\textbf{GAN-based Methods.}
GANs~\cite{goodfellow2014generative} have proven to be effective in generative tasks by considering global consistency instead of local, as well as pixel level consistency. This property makes adversarial learning suitable for unsupervised depth estimation, allowing the generation of more accurate viewpoint images. Pilzer~\etal~\cite{pilzer2018unsupervised} propose to learn a stereo matching model in a cycled, adversarial fashion, while~\cite{adadepth} utilize it in a context of domain adaptation for a single-track network, using a semi-supervised setting with additional synthetic data. In this work, we use adversarial learning at the generator level, however, we present a novel dual GAN design and use a CRF model to couple the network for a structured refinement and fusion of both generator and discriminator outputs in an end-to-end fashion.

\section{The Proposed Approach}
\vspace{-5pt}
In this section, we present the proposed approach for unsupervised monocular depth estimation. A framework overview is depicted in Fig.~\ref{framework}. We first introduce the designed dual generative adversarial network, and then elaborate how we couple the two sub-networks upon both the generator and the discriminator, and perform a structured refinement of the outputs within a joint CRF model. Finally, we describe how the whole model can be organized into a unified deep network and can be simultaneously optimized in an end-to-end fashion. 
\vspace{-5pt}
\subsection{Dual Generative Adversarial Networks}
\vspace{-3pt}
\par\noindent\textbf{Basic Network Structure.} As formalized in previous works~\cite{godard2016unsupervised,zhan2018unsupervised}, unsupervised monocular depth estimation can be treated as a problem of learning a dense correspondence field between two calibrated image spaces. Given a set of $N$ stereo image pairs $\{(\vect{I}_{l}^n, \vect{I}_{r}^n)\}_{n=1}^N$, the target is to learn a 
generator $G$ which is able to estimate the dense correspondence (\ie~the disparity map) $\vect{d}_{r}^n$ from $\vect{I}_{l}^n$ to $\vect{I}_{r}^n$, and the supervision is obtained from a reconstruction of $\vect{I}_{r}^n$ using a warping function $f_w$, \ie~$\vect{\tilde{I}}_{r}^n = f_w(\vect{d}_{r}^n, \vect{I}_{l}^n)$. The network can be optimized by minimizing the difference between $\vect{I}_r^n$ and $\vect{\tilde{I}}_r^n$. As shown in Fig.~\ref{framework}, we propose a dual generative adversarial network with a pair of stereo images $(\vect{I}_{l}^n, \vect{I}_{r}^n)$ as input in the training phase. The two generative networks $G_a$ and $G_b$ are designed to estimate two disparity maps $\vect{d}_{r_a}^n$ and $\vect{d}_{r_b}^n$ respectively,
and part of shallow layers of $G_a$ and $G_b$ are shared to reduce the network capacity. Then two warping functions $f_{w_a}$ and $f_{w_b}$ are separately used to generate two synthesized right-view images via sampling from the same left-view image $\vect{I}_{l}^n$. Since $\vect{d}_{r_a}^n$ and $\vect{d}_{r_b}^n$ are generated from different inputs while similar images and the warping is performed on the the same image, the two disparity maps are well aligned and are complementary to each other. For the synthesized images, we use two discriminators $D_a$ and $D_b$ to benefit from the advantage of adversarial learning. To only learn the dual generative adversarial network, the optimization objective is:
\begin{equation}
  \begin{split}
  \hspace*{-0.3cm} &\!\!\mathcal{L}_{gan}(G_a, G_b, D_a, D_b, \vect{I}_{l}^n, \vect{I}_{r}^n)= \\
 \!\! &\!\!\mathbb{E}_{\vect{I}_r^n \sim p(\vect{I}_r^n)}[\log D_b(\vect{I}_r^n)] \!\! +\!\!  \mathbb{E}_{\vect{I}_l^n \sim p(\vect{I}_l^n)}[\log(1 \!\!-\!\! D_b(G_a(\vect{I}_l^n)))] + 
  \\ \!\! &\!\!\mathbb{E}_{\vect{I}_r^n \sim p(\vect{I}_r^n)}[\log D_a(\vect{I}_r^n)] \!\! +\!\!   \mathbb{E}_{\vect{I}_r^n \sim p(\vect{I}_r^n)}[\log(1 \!\!-\!\! D_a(G_b(\vect{I}_r^n)))]\!\!
  \end{split}
  \label{newGANLoss}
\end{equation}
We adopt a sigmoid cross entropy to measure the expectation of the image $\vect{I}_l$ and $\vect{I}_r$ against the distribution $p(\vect{I}_l)$ and $p(\vect{I}_r)$ of the left- and right-view images respectively. Along with the adversarial objective, we have also an $L_1$ reconstruction objective for the generators:
\begin{equation}
	\mathcal{L}_{rec} (G_a, G_b, \vect{I}_{r}^n, \vect{I}_{l}^n) = \parallel \tilde{\vect{{I}}}_{r}^n - \vect{I}_{r}^n\parallel_1 + \parallel \tilde{\vect{{I}}}_{l}^n - \vect{I}_{r}^n\parallel_1
\end{equation}
where $\tilde{\vect{{I}}}_{l}^n = f_w(\vect{d}_{r}^n, \vect{I}_{l}^n)$ and $\tilde{\vect{{I}}}_{r}^n = f_w(\vect{d}_{l}^n, \vect{I}_{r}^n)$ are the synthesized images with the disparity maps $\vect{d}_{r}^n$ and $\vect{d}_{l}^n$ estimated by the two generators $G_a$ and $G_b$ respectively.

\par\noindent\textbf{Network Hallucination.} Monocular depth estimation uses only a single image as input in the test phase. To achieve this, we designed a hallucination sub-network $H(\cdot) $ with a convolutional encoder-decoder structure, which aims at approximating the disparity map $\vect{d}_{r_b}^n$ using $\vect{d}_{r_a}^n$, \ie~$\vect{d}_{r_h}^n = H(\vect{d}_{r_a}^n, \vect{W}_h)$, where $\vect{W}_h$ are the parameters of the network $H$. In this way, the network $H$ preserves the information coming from the image $\vect{I}_r^n$, while only the input image $\vect{I}_l^i$ is required in the testing. During the training we use an $L_1$ loss to optimize the network parameters $\vect{W}_h$ as follows: $ \mathcal{L}_h(\vect{d}_{r_a}^n, \vect{d}_{r_b}^n, \vect{W}_h) = \sum_{n=1}^N \parallel H(\vect{d}_{r_a}^n, \vect{W}_h) - \vect{d}_{r_b}^n \parallel_1$. The proposed approach is general, if we have the stereo images in the testing phase, the network $H$ can be disabled to support testing with stereo images.
\vspace{-5pt}
\subsection{Structured Coupling via Deep CRFs}
\vspace{-3pt}
Probabilistic graphical models such as conditional random fields (CRFs) have shown great success in supervised-based approaches~\cite{liu2016learningtpami,xu2017multi}. We investigate here how the CRF can be used for structured unsupervised monocular depth estimation. Since we have two generative adversarial networks, we propose a CRF coupling model for a structured fusion of the outputs of the two nets from both the generator and the discriminator. We first give the formulation of our model in coupling two disparity maps from the two generators, and then illustrate how this can also be done together with the two adversarial score maps.

Given the observed disparity maps $\vect{d}_{r_a}$ and $\vect{d}_{r_h}$ from the backbone network, let us denote $\vect{d}_r$ as a hidden disparity map to be inferred, and $d_{r}^i$ is an element of $\vect{d}_{r}$ at position $i$ (in analogy to $\vect{d}_{r_h}$ and $\vect{d}_{r_a}$). The model can be expressed as a Gibbs conditional distribution $P(\vect{d}_r | \vect{d}_{r_a}, \vect{d}_{r_h}, \vect{I}_r, \vect{\Theta}) = \exp (-E(\vect{d}_{r} | \vect{d}_{r_a}, \vect{d}_{r_h}, \vect{I}_r, \vect{\Theta}))/Z(\vect{I}_r, \vect{\Theta})$, where $\vect{\Theta}$ is a set of parameters and, $E$ and $Z$ are an energy and a normalization function, respectively. We formally define the energy in Eq. \ref{eq:energy}, where $\vect{f}_i$ and $\vect{f}_j$ are features calculated from the input image $\vect{I}_r$ at position $i$ and $j$; $\alpha_1>0$ and $\alpha_2>0$ are weighting factors for the two unary terms; $k_l$ is a gaussian kernel for similarity between the features:
\begin{equation}
 \setlength{\belowdisplayskip}{0pt}
\begin{split}
E(\vect{d}_{r} | \vect{d}_{r_a}, & \vect{d}_{r_h}, \vect{I}_r, \vect{\Theta})) =  \\ &\sum_i \big(\alpha_1 (d_{r}^i - d_{r_a}^i)^2 + \alpha_2 (d_{r}^i - d_{r_h}^i)^2\big) \\
 + & \sum_{i \neq j}\sum_l\big(\beta_l k_l(\vect{f}_i^{(l)}, \vect{f}_j^{(l)})(d_r^{i}-d_r^{j})^2\big),
 \end{split}
 \setlength{\belowdisplayskip}{5pt}
 \label{eq:energy}
\end{equation}
For the unary term, an isotropic Gaussian function is used to describe the potential between the observation and the hidden disparity map, as a constrain that the hidden map to be as close as possible to the observation ones. For the pairwise term, following~\cite{koltun2011efficient} we use both an appearance and a smoothness kernel (\ie for~$l$=1, 2 then $\beta_l$ are weights for the kernels) to have structured constraints on the hidden disparity map.

\par\noindent\textbf{Inference.}~Exact inference of the fully connected model requires high complexity because of the calculation of inverse matrices~\cite{liu2016learningtpami,ristovski2013continuous}. We approximate the inference using mean-field theory. The target is to approximate the distribution $P(\vect{d}_r | \vect{d}_{r_a}, \vect{d}_{r_h})$ with another simpler distribution $Q(\vect{d}_r | \vect{d}_{r_a}, \vect{d}_{r_h})$ which can be expressed with a set of independent marginal distributions, \ie~$Q(\vect{d}_r | \vect{d}_{r_a}, \vect{d}_{r_h}) = \prod_{i} Q_i(d_r^i | \vect{d}_{r_a}, \vect{d}_{r_h})$. We obtain an optimal solution $\tilde{Q}$ by minimizing the KL divergence between the distribution $P$ and $Q$, \ie~$\log\tilde{Q}_i(d_r^i | \vect{d}_{r_a}, \vect{d}_{r_h})=\mathbb{E}_{j\neq i}[\log P(\vect{d}_r | \vect{d}_{r_a}, \vect{d}_{r_h})] + \rm{C}$ with $\rm{C}$ as a const. The mean-field inference for $Q$ can be derived as follows: 
\begin{multline}
\setlength{\belowdisplayskip}{0pt}
\tilde{Q}_i(d_r^i) \propto \exp \big(-(\alpha_1+ \alpha_2) {d_r^i}^2 + 2d_r^i(\alpha_1 d_{r_a}^i + \alpha_2 d_{r_h}^i)  \\ - \sum_{l}\beta_l k_l(\vect{f}_i^{(l)}, \vect{f}_j^{(l)}) ({d_r^i}^2 - 2 d_r^i d_r^i])\big).
\setlength{\belowdisplayskip}{0pt}
\end{multline}
The equation implies that the log distribution of $\tilde{Q}_i$ takes a Gaussian distribution and its expectation produces the maximum probability. Then we have the mean-field updating for the continuous hidden variable $d_r^i$ written as
\begin{equation}
d_r^i = \frac{\alpha_1 d_{r_a}^i + \alpha_2 d_{r_h}^i + \sum_{l}\sum_{j\neq i} k_l(\vect{f}_i^{(l)}, \vect{f}_j^{(l)}) d_r^j}{\alpha_1 + \alpha_2 + \sum_{l} \sum_{j\neq i} k_l(\vect{f}_i^{(l)}, \vect{f}_j^{(l)})}
\label{mean-field}
\end{equation}
The updating of $d_r^i$ is an iterative operation, and we are able to achieve a local minimum after $T$ iterations. In the following we discuss how we implemented the continuous mean-field in neural network (NMF) for the inference of the hidden variables, enabling a joint end-to-end optimization with the proposed backbone dual GAN network.

\par\noindent\textbf{Mean-Field Updating in Neural Network (NMF).} In Eq.~\ref{mean-field}, we have three steps to perform the mean-field updating. The first step is a linear combination of the unary terms, \ie~$\alpha_1 d_{r_a}^i + \alpha_2 d_{r_h}^i$, which can be implemented with $1\times 1$ convolutions with a ReLU operation, and then an element-wise addition operation. The second step is the message passing. To calculate the message with the Gaussian convolution operation, \ie~$\sum_{j\neq i} k_l(\vect{f}_i^{(l)}, \vect{f}_j^{(l)}) d_r^j$, due to the high complexity, we utilize a local receptive field considering a locally connected graph. The message passing can be performed using element-wise addition operation. The third step is a normalization step. The calculation of the normalization factor (\ie~the denominator) is similar to that of the previous steps, and an element-wise division operation is used to perform the normalization. We have in total four parameters to optimize, \ie~two linear combination weights for the observation maps $d_{r_a}$ and $d_{r_h}$, and other two weights for the gaussian kernels. Since each forward step is differentiable, the mean-field updating can be optimized with the back-propagation, and we can stack several mean-field blocks by sharing parameters for a deep CRF inference. 

\par
\begin{figure}[!t]
\centering
\includegraphics[width=0.99\linewidth]{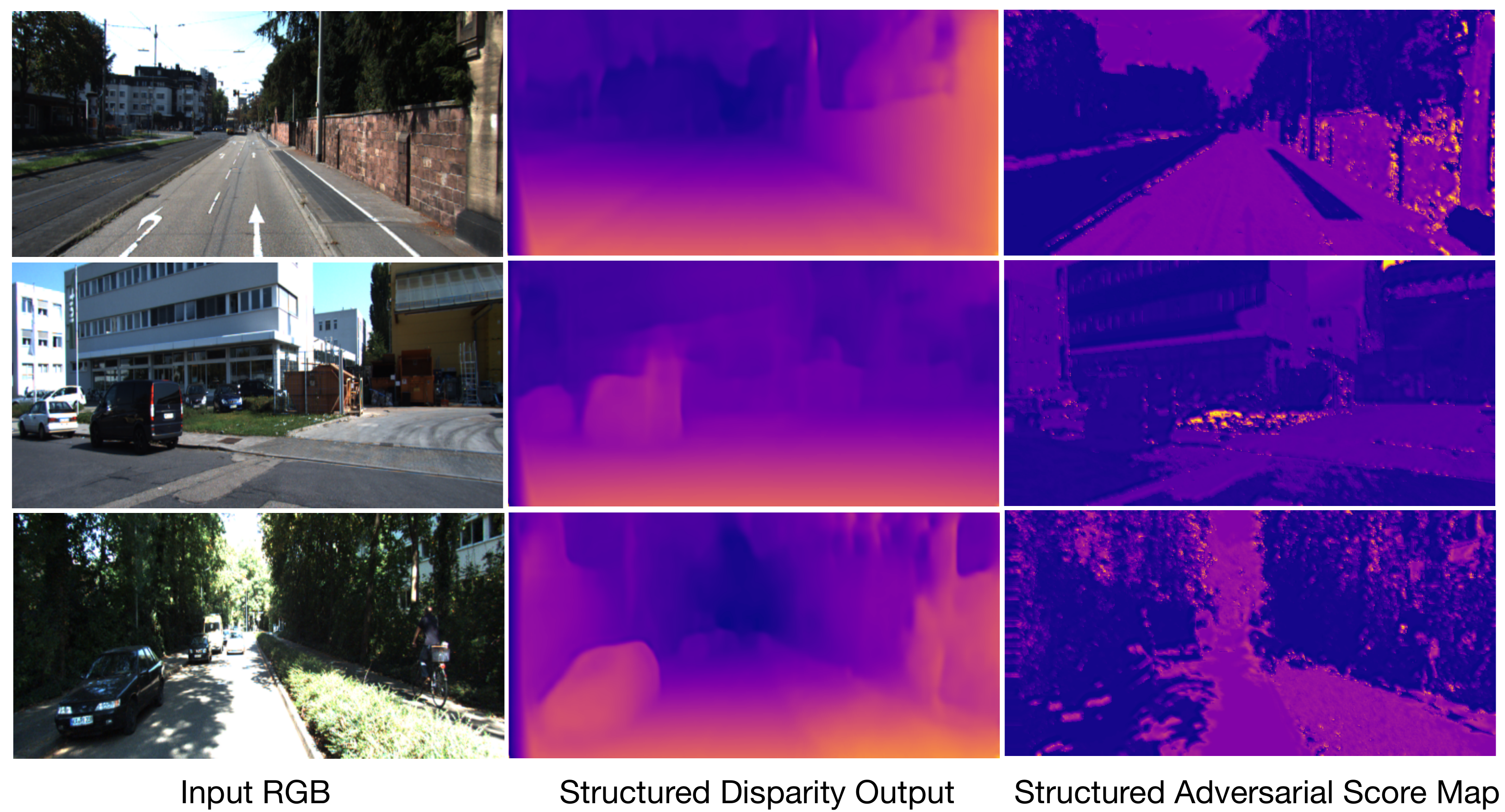} 
\vspace{-0mm}
\caption{Examples of the structured output of the disparity maps and the adversarial score maps on KITTI using the proposed CRF-DGAN. The CRF model couples not only two GAN sub-networks but also connects the generators and the discriminators with mutual constraints in joint optimization.}
\label{motivation2}
\vspace{-0pt}
\end{figure}

\noindent\textbf{Joint Coupling of the Generator and Discriminator.} To model the structured relationship between the generator and discriminator, we use one single CRF model to learn the fusion and refinement of both. The discriminators and the generators from the dual GAN produce the same number of outputs, \ie~two disparity maps and correspondingly two real/fake adversarial score maps, where we consider a pixel-level discriminator. Then we respectively input them into the deep CRF coupling model introduced above with two separate forward computations, and collect gradients from both to perform one backward computation to update the model parameters during learning. By doing so, the outputs from the generators and the discriminators will jointly affect the model learning, contributing implicitly as mutual constraints to better optimize both parts. Fig.~\ref{motivation2} shows examples of structured outputs of the generated disparity and the discriminative errors. We use only one adversarial loss using the refined and fused the adversarial score map from the deep coupling CRF model. Let us denote $D_{\mathrm{real}}^{\mathrm{crf}}$ and $D_{\mathrm{fake}}^{\mathrm{crf}}$ as the adversarial score for the real and fake samples, and thus we replace Eq.~\ref{newGANLoss} as:
\begin{equation}
  \begin{split}
  \hspace*{-0.3cm} 
  &\mathcal{L}_{gan}^{\mathrm{crf}}(G_a, G_b, D_a, D_b, \vect{I}_{l}^n, \vect{I}_{r}^n) = \\ 
  & \mathbb{E}_{\vect{I}_r^n \sim p(\vect{I}_{r}^n) }[\log D_{\mathrm{real}}^{\mathrm{crf}}] + \mathbb{E}_{\vect{I}_l^n,\vect{I}_{r}^n \sim p(\vect{I}_l^n, \vect{I}_{r}^n)}[\log(1 - D_{\mathrm{fake}}^{\mathrm{crf}})].
  \hspace*{-0.3cm} 
  \end{split}
\end{equation}


\par\noindent\textbf{End-to-End Joint Optimization.} The learning of the whole network involves optimization of both the dual generative adversarial network and the deep CRF model. For the CRF model, the final output disparity map is used to synthesize another right-view image $\vect{d}_{rc}^n$, and we use an $l_1$ reconstruction loss $\mathcal{L}_{crf}$ to supervise the learning of the CRF model with $\mathcal{L}_{crf} = \sum_{n=1}^N || \vect{I}_{rc,n} - \vect{I}_{r,n} ||_1$. To combine the loss functions of the dual generative adversarial network, the whole deep network optimization objective becomes: $\mathcal{L}_o = \gamma_1 \mathcal{L}_{rec} + \gamma_2 \mathcal{L}_h + \gamma_3 (\mathcal{L}_{gan}^{\mathrm{crf}} + \mathcal{L}_{crf})$, where $\{\gamma_i\}_{i=1}^3$ is a set of weights for balancing the loss from different parts.

\section{Experiments}
\begin{figure*}[tbp]
\centering
\includegraphics[width=1\linewidth]{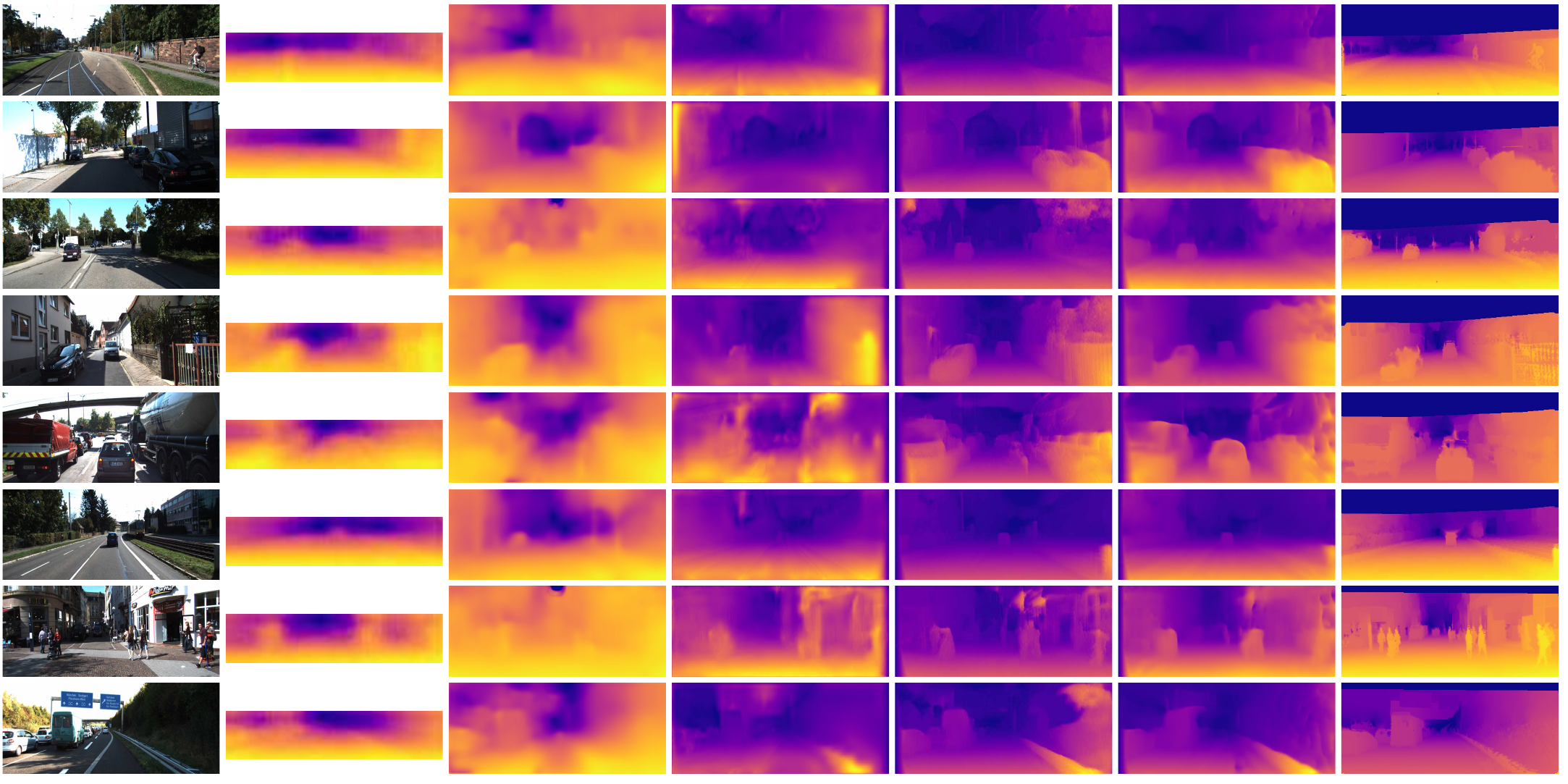} 
\put(-480,255){\scriptsize RGB Image}
\put(-412,255){\scriptsize Eigen~\etal~\cite{eigen2014depth}}
\put(-337,255){\scriptsize Zhou~\etal~\cite{zhou2017unsupervised}}
\put(-273,255){\scriptsize Garg~\etal~\cite{garg2016unsupervised}}
\put(-202,255){\scriptsize Godard~\etal~\cite{godard2016unsupervised}}
\put(-116,255){\scriptsize Ours }
\put(-59,255){\scriptsize GT Depth Map}
\vspace*{-0mm}
\caption{Examples of depth prediction results on the KITTI raw dataset. Qualitative comparison with other depth estimation methods on this dataset is presented. The sparse ground-truth depth maps are interpolated for better visualization.}
\label{KITTI_comp}
\vspace{-5pt}
\end{figure*}
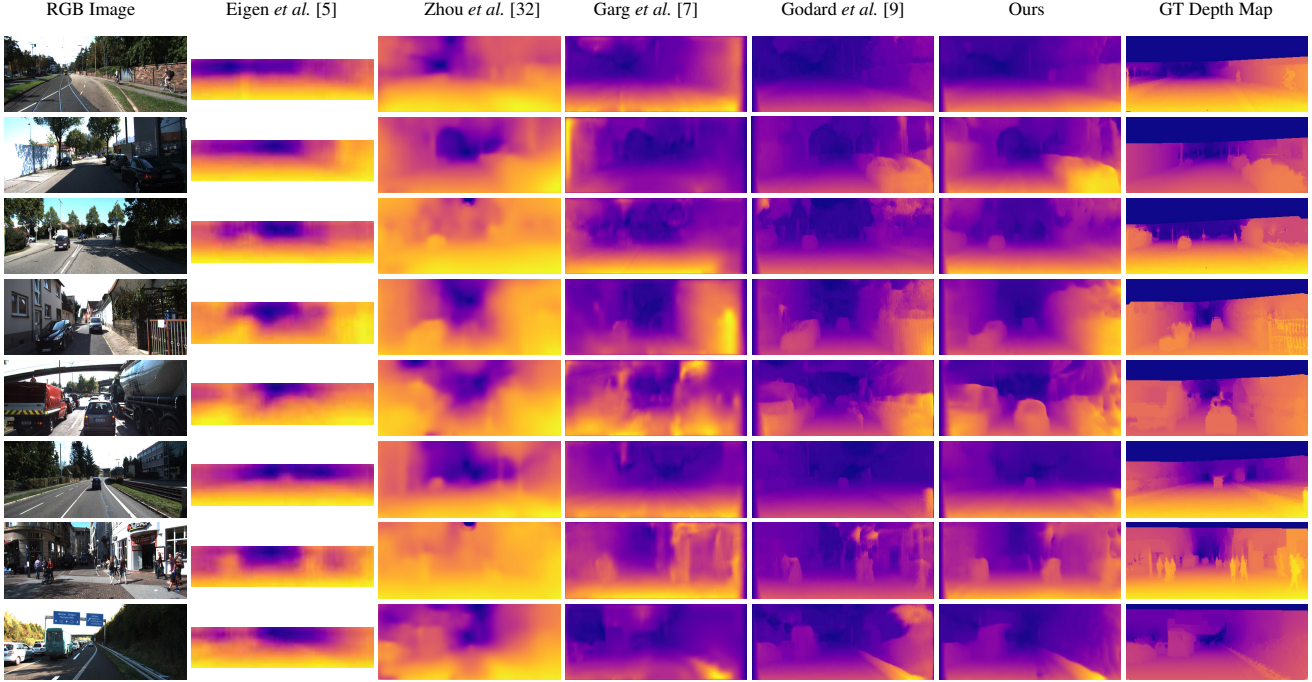

\begin{table*}[!t]
\centering
\Huge
\setlength\tabcolsep{10pt}
\begin{adjustbox}{width=0.92\linewidth}

\begin{tabular}{l|cccc|ccc}
\toprule[2.5pt]
\multirow{2}{*}{\tabincell{c}{Method}} & \multicolumn{4}{c|}{\tabincell{c}{Error  (lower is better)}} & \multicolumn{3}{c}{\tabincell{c}{Accuracy  (higher is better)}} \\\cline{2-7}
                                      & rel & sq rel & rms & rms log & $\delta < 1.25$ & $\delta < 1.25^2$ & $\delta < 1.25^3$ \\\midrule\midrule

CRF-DGAN (baseline model)        & 0.1650 &1.7563 &6.164 & -&0.773 &0.914 &0.962 \\
CRF-DGAN (w/ deep network hallucination ) &0.1617 &1.4834 &5.991 &0.242 &0.779 &0.917 &0.964 \\
CRF-DGAN (w/ adversarial learning ) &0.1528 &1.4005 &6.029 &0.247 &0.785 &0.918 &0.965 \\
CRF-DGAN (w/ coupled adversarial learning) &0.1423 &1.3067 &5.687 &0.238 &0.813 &0.928 &0.968 \\
CRF-DGAN (w/ dual coupled adversarial learning) &\textbf{0.1407} &\textbf{1.2831} & \textbf{5.677} & \textbf{0.237} & \textbf{0.815} &\textbf{ 0.930} & \textbf{0.968} \\
\bottomrule[2.5pt]                            
\end{tabular}
\end{adjustbox}
\vspace{-0mm}
\caption{Quantitative analysis of the main components of our method on the KITTI dataset. The evaluation is conducted on the predicted depth maps following the standard evaluation protocol. 
}

\label{kitti_ablation}
\vspace{-5pt}
\end{table*}
\vspace{-0mm}

\begin{table*}[!h]
\centering
\Huge
\setlength\tabcolsep{10pt}
\begin{adjustbox}{width=0.92\linewidth}
\begin{tabular}{l|cccc|ccc}
	\toprule[2.5pt]
	\multirow{2}{*}{\tabincell{c}{Method}} & \multicolumn{4}{c|}{\tabincell{c}{Error  (lower is better)}} & \multicolumn{3}{c}{\tabincell{c}{Accuracy  (higher is better)}} \\\cline{2-8}
	& rel & sq rel& rms & rms log & $\delta < 1.25$ & $\delta < 1.25^2$ & $\delta < 1.25^3$ \\\midrule\midrule
	CRF-DGAN (baseline model)       & 0.4676 & 7.3992 & 5.741 &0.493 & 0.735 & 0.890  & 0.945\\ 
	CRF-DGAN (w/ deep network hallucination )         &  0.4397 & 6.3369 &  5.444 &0.456 & 0.730  &  0.887 & 0.944 \\
	CRF-DGAN (w/ adversarial learning )   &  0.4327 & 6.2006 &  5.541 &0.424 & 0.738  &  0.890 & 0.944  \\
	CRF-DGAN (w/ coupled adversarial learning)  & \textbf{0.4109}  & \textbf{5.9848} & \textbf{4.636} & \textbf{0.403} & \textbf{0.756} & \textbf{0.897} & \textbf{0.953} \\
\bottomrule[2.5pt]                            
\end{tabular}
\end{adjustbox}
\vspace{-0mm}
\caption{Quantitative analysis of the main components of our method on the Cityscapes dataset. Cityscapes does not provide a standard evaluation protocol for depth estimation. We directly evaluate the performance on the predicted disparity maps.}

\label{cityscape_ablation}
\vspace{-5pt}
\end{table*}

\vspace{-5pt}
We now present the experimental setup and results to demonstrate the effectiveness of the proposed approach.
\vspace{-5pt}
\subsection{Experimental Setup}
\vspace{-3pt}
\noindent\textbf{Datasets.} We have conducted experiments on the KITTI~\cite{Geiger2013IJRR}, Cityscapes~\cite{Cordts2016Cityscapes} and Make3D~\cite{saxena2006learning,saxena2009make3d} datasets. 
The \textbf{KITTI} dataset contains depth images captured with a LiDAR sensor mounted on a driving vehicle. In our experiments we follow the experimental protocol proposed by~\cite{eigen2014depth} containing 22,600 training images and 697 images test images. The RGB image resolution is reduced by half with respect to the original $1224\times 368$ pixels. To evaluate the transfer learning capabilities of our method, we test the model trained on Cityscapes and evaluate it on the \textbf{Make3D} dataset, which contains only 400 single training RGB and depth map pairs, and 134 test samples. The \textbf{Cityscapes} is a large-scale dataset mainly used for semantic urban scene understanding. The annotated split contains 2975 training, 500 validation, and 1525 test images. The dataset also provides pre-computed disparity maps associated with the rgb images. As the images of the dataset have a high resolution ($2048 \times 1024$), we resize the image to size of $512 \times 256$ as in~\cite{godard2016unsupervised} for training due to the limitation of the GPU memory, and the bottom one fifth of the image is removed.

\noindent\textbf{Evaluation Metrics.} Following~\cite{eigen2015predicting,eigen2014depth,wang2015towards}, we consider several evaluation metrics to quantitatively assess the performance of our approach: the mean relative error (rel), root mean squared error (rms), mean log10 error (log10), and a thresholded accuracy.

Specifically if $Q$ is the total number of pixels of the test set and $\bar{d}_i$ and $d_i$ denote the estimated 
and the ground-truth depth for pixel $i$, we compute:
(i) the mean relative error (rel): 
\( \frac{1}{Q}\sum_{i=1}^Q\frac{|\bar{d}_i - d_i|}{d_i} \); (ii) the
root mean squared error (rms): 
\( \sqrt{\frac{1}{Q}\sum_{i=1}^Q(\bar{d}_i - d_i)^2} \); (iii) the
mean log10 error (log10): 
\( \frac{1}{Q}\sum_{i=1}^Q \Vert \log_{10}(\bar{d}_i) - \log_{10}(d_i) \Vert \) and (iv) the 
accuracy with threshold $t$, \ie the percentage of $\bar{d}_i$ such that $\delta=\max (\frac{d_i}{\bar{d}_i}, \frac{\bar{d}_i}{d_i}) < t$, 
where $t \in [1.25, 1.25^2, 1.25^3]$. 
In order to compare our results with previous methods on the KITTI dataset we crop our images using the evaluation crop applied by  \cite{eigen2014depth}.

\noindent\textbf{Implementation Details.}
Messages are passed via locally connected convolutions \ie~considering a local receptive field for the Gaussian convolution with a kernel window size of $15\times15$. In our CRF model we consider dependencies only for the last scale. The initial learning rate is set to 1e-4 in all our experiments, and decreases 5 times after for each step reached in $[30000, 55000]$. The momentum and weight decay parameters are set to 0.9 and 0.0002, as in~\cite{xie2015holistically}. The batch size of the algorithm is set to 8.

\begin{figure*}[tbp]
\centering
\vspace{-25mm}
\includegraphics[angle=270, width=0.99\linewidth]{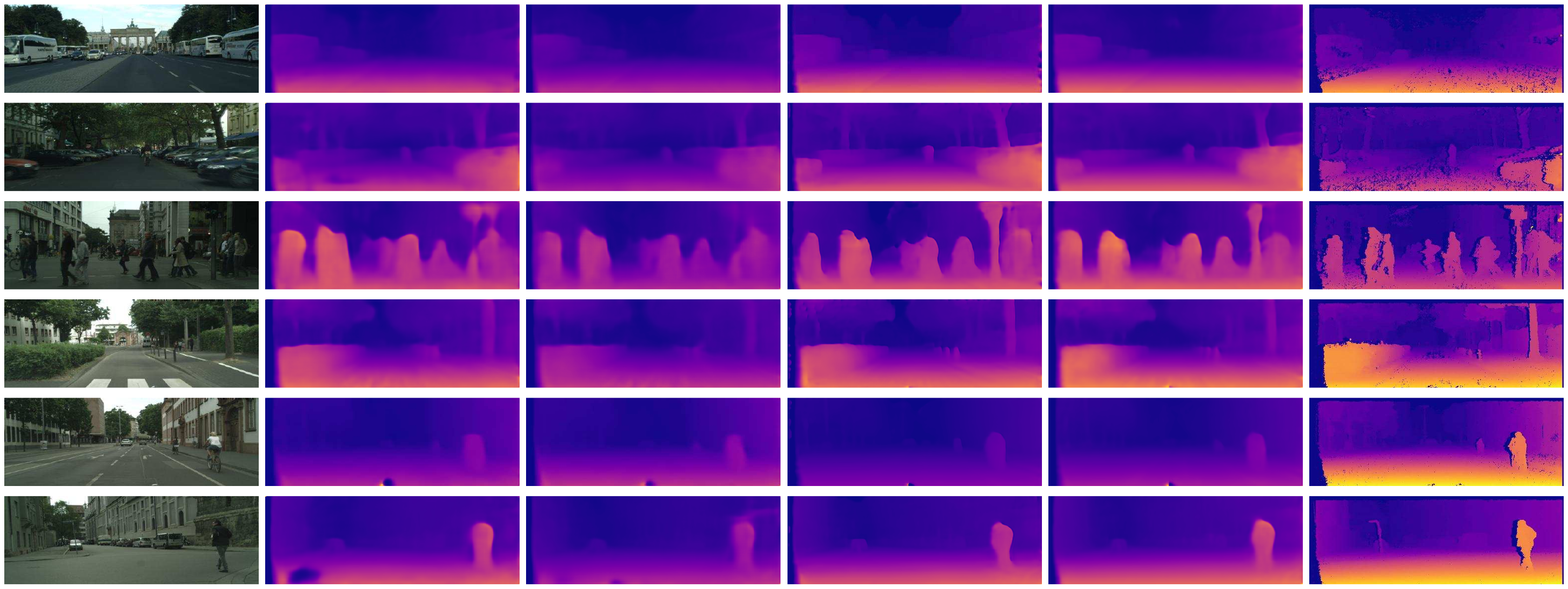}
\put(-470,-75){\scriptsize RGB Image}
\put(-385,-75){\scriptsize Baseline (i)}
\put(-305,-75){\scriptsize Baseline (ii)}
\put(-220,-75){\scriptsize Baseline (iii)}
\put(-135,-75){\scriptsize Full model}
\put(-58,-75){\scriptsize GT Depth Map} 
\vspace{-26mm}
\caption{Qualitative comparison of different variants of the proposed CRF-DGAN model on the Cityscapes dataset.}
\label{Cityscapes_comp}
\end{figure*}

\begin{table*}[!t]
\centering
\setlength\tabcolsep{8pt}
\begin{adjustbox}{width=0.95\linewidth}
\begin{tabular}{l|cc|cccc|ccc}
\toprule[1.3pt]  
\multirow{2}{*}{Method} & \multicolumn{2}{c|}{Setting} 
& \multicolumn{4}{c|}{\tabincell{c}{Error (lower is better)}} & \multicolumn{3}{c}{\tabincell{c}{Accuracy (higher is better)}} \\\cline{2-10}
                                      & cap & supervised? 
                                      & rel & sq rel & RMSE & RMSE (log) & $\delta < 1.25$ & $\delta < 1.25^2$ & $\delta < 1.25^3$ \\\midrule\midrule
Saxena~\etal~\cite{saxena2009make3d}   & 80m & $\surd$ 
& 0.280  &  - & 8.734 &  & 0.601  &  0.820  &  0.926\\
Eigen~\etal~\cite{eigen2014depth}    & 80m & $\surd$ 
& 0.203  &  1.548   &  6.307 & 0.282 &  0.702 &  0.890  &  0.958 \\
Liu~\etal~\cite{liu2016learningtpami}    & 80m & $\surd$ 
& 0.202  &  1.614   &  6.523 & 0.275 &  0.678 &  0.895  &  0.965 \\
AdaDepth~\cite{adadepth}* & 50m & $\surd$  & 0.162 & 1.041 & 4.344 & 0.225  & 0.784 & 0.930 & 0.974 \\ 
Kuznietsov~\etal~\cite{kuznietsov2017semi} & 80m & $\surd$ & - & - & 4.815 & 0.194  & {0.845} & {0.957} & {0.987} \\
Xu~\etal~\cite{xu2018structured} & 80m & $\surd$ & 0.120 & 0.764 & 4.341 & 0.181 & 0.852 & 0.959 & 0.987 \\
Gan~\etal~\cite{gan2018monocular}    & 80m & $\surd$ 
& 0.098  &  0.666  &  3.933 & 0.173 &  0.890 &  0.964  &  0.985 \\
\midrule\midrule
Garg~\etal~\cite{garg2016unsupervised}  & 80m & \ding{53} 
& 0.177 & 1.169  & 5.285    & 0.282  &  0.727 & 0.896  &  0.962   \\
Garg~\etal~\cite{garg2016unsupervised} L12 + Aug 8x  & 50m & \ding{53} 
& 0.169  & 1.080   & 5.104    & 0.273  &  0.740 & 0.904  &  0.958   \\
Godard~\etal~\cite{godard2016unsupervised}   & 80m & \ding{53}
& 0.148 & 1.344  & 5.927    &  0.247 &  0.803 & 0.922  &  0.963   \\
Kuznietsov~\etal~\cite{kuznietsov2017semi} & 80m & \ding{53}  & - & - & {8.700} & 0.367 & {0.752} & {0.904} & {0.952} \\ 
Zhou~\etal~\cite{zhou2017unsupervised} & 80m & \ding{53} 
& 0.208   &  1.768  & 6.858  &  0.283 & 0.678   & 0.885   & 0.957  \\
AdaDepth~\cite{adadepth} & 50m & \ding{53}  & 0.203 & 1.734 & 6.251 & 0.284 & 0.687 & 0.899 & 0.958 \\ 
Mahjourian~\etal~\cite{mahjourian2018unsupervised}$\dagger$  & 80m & \ding{53} & 0.163 & 1.240 & 6.220 &  0.250 & 0.762 & 0.916 & 0.968 \\
Pilzer~\etal~\cite{pilzer2018unsupervised} & 80m & \ding{53} & 0.152 & 1.388 & 6.016 & 0.247  & 0.789 & 0.918 & 0.965 \\
Wang~\etal~\cite{wang2017learning} & 80m & \ding{53} & 0.151 & 1.257 & 5.583 & \textbf{0.228}  & 0.810 & \textbf{0.936} & \textbf{0.974} \\ 
Zou~\etal~\cite{zou2018df}$\dagger$ & 80m & \ding{53} & 0.150 & 1.124 & 5.507 & 0.223 & 0.806 & 0.933 & 0.973 \\
Zhan~\etal~\cite{zhan2018unsupervised}$\dagger$  & 80m & \ding{53} & 0.144 & 1.391 & 5.869 & 0.241  & 0.803 & 0.933 & 0.971 \\
Guo~\etal~\cite{guo2018learning}* & 80m & \ding{53} & 0.105 & 0.811 & 4.634 & 0.189  & 0.874 & 0.959 & 0.982 \\
 \midrule\midrule

CRF-DGAN (ours) & 80m & \ding{53} &\textbf{0.1354} & \textbf{1.1815} & \textbf{5.582} & 0.235 & \textbf{0.828} & 0.933 & 0.967 \\ 
CRF-DGAN (ours) & 50m & \ding{53} & \textbf{0.1283} & \textbf{0.8681} & \textbf{4.223} & \textbf{0.222} & \textbf{0.840} & \textbf{0.941} & \textbf{0.971}\\

\bottomrule[1.3pt]                             
\end{tabular}
\end{adjustbox}
\vspace{2mm}
\caption{State of the art comparison on the KITTI dataset. Methods that require additional image data are marked with *, and those that require video data are marked with $\dagger$. We bold the metrics where our method achieves the best results under the same settings. }
\vspace{2mm}
\label{sota_KITTI}
\vspace{-5mm}
\end{table*}

\vspace{-5pt}
\subsection{Experimental Results}
\vspace{-3pt}
We first conduct an in-depth analysis of the proposed approach, and then carry out a state-of-the-art comparison with other competing methods, and finally provide a discussion on the qualitative results.

\noindent\textbf{Baseline Models.} 
We mainly aim to demonstrate the effectiveness of the proposed approach from three aspect: first, the monocular depth estimation with adversarial learning strategy, second, the proposed dual GAN network structure, and third, the coupling scheme to fuse and refine the proposed dual GAN in a structured fashion.  Thus we present an ablation study based on several baselines, including \textbf{(i)} CRF-DGAN (baseline model): a single branch model which uses only the generator without using the adversarial loss; \textbf{(ii)} CRF-DGAN (w/ deep network hallucination): a dual-branch model with network hallucination, which has two branches each synthesizing a right view new image, and sharing the parameters of the encoder part. The dual-branch model is used as the backbone network structure of our approach. A hallucinator is added in order to predict images in a monocular fashion in the testing phase; \textbf{(iii)} CRF-DGAN (w/ adversarial learning): we train the backbone adversarially, \ie~adding a discriminator per branch;  \textbf{(iv)} CRF-DGAN (w/ coupled adversarial learning): the two discriminators of the dual-GAN are coupled with the proposed CRF model; \textbf{(v)} CRF-DGAN (w/ dual coupled adversarial learning): both the discriminators and the generators are coupled with the proposed CRF model. 

\noindent\textbf{Model Analysis.} We conduct the ablation study on the KITTI raw and Cityscapes datasets, as shown in Table~\ref{kitti_ablation} and~\ref{cityscape_ablation}. Comparing baseline (i) and (ii), we observe a minor improvement in absolute error, but a more substantial improvement in all accuracy metrics, especially on Cityscapes dataset. This performance boost is likely caused by the network hallucination learning the complimentary information between the two stereo viewpoints, resulting in a better learned model. The effectiveness of adversarial learning has been demonstrated in other GAN-based monocular depth estimation works~\cite{adadepth,pilzer2018unsupervised}, a benefit also observed between the baseline models (iii) and (ii). Baseline models (iv) and (v) evaluate the effectiveness of the proposed CRF model using different coupling strategies, between the disparity maps produced by the generators and the adversarial score maps by the discriminators. By comparing (v) and (iii), we have 1.2 points gain on the metric rel on KITTI. We should note that this is not a trivial gain on this very challenging and almost performance-saturated dataset. From the accuracy aspects: we improve 4 points from 0.779 to 0.813, clearly demonstrating the effectiveness of the proposed CRF-based structured coupling approach. We observe a more significant boost (around 2.7 points) on the rel metric on the Cityscapes dataset, and that coupling both the discriminators and the generators achieves better performance than coupling only the discriminators, meaning that the joint coupling brings extra constraints for each part, and facilitates the network optimization, confirming our initial motivation. In overall, our approach has 2.5 points gain on the difficult rel metric over the single branch basic baseline model, demonstrating the effectiveness of the CRF coupled dual GAN structure.


\vspace{0mm}
\noindent\textbf{State-of-the-art Comparison.} 
Table~\ref{sota_KITTI} compares the depth estimation results of our full CRF-DGAN model with other supervised and unsupervised methods. We outperform most competitors due to a joint structured optimization of both the discriminators and the generators sign. The concise network design also facilitates the overall optimization process. With regard to \cite{garg2016unsupervised}, we also report results for a 50m depth cap. The full CRF-DGAN model achieves better performance in both the 50m and 80m settings. Of interest is that CRF-DGAN outperforms \cite{godard2016unsupervised} in all metrics. Our performance is much better than AdaDepth which also considers generative adversarial networks while used extra synthetic training data. \cite{zhan2018unsupervised}, DF-Net~\cite{zou2018df}, and \cite{mahjourian2018unsupervised} do not use the same setting as our approach, requiring video training data for extra temporal information. In contrast, CRF-DGAN requires only image pairs in training and single images in testing. Although our approach is not directly comparable to them,  it outperforms their results on all the metrics. Our approach is outperformed by \cite{gan2018monocular}, which uses stereo-matching techniques to improve upon available sparse LiDaR ground truth. As so it is a method with a different setting to ours and is not directly comparable. \cite{guo2018learning} uses a stereo model trained on the KITTI raw and then synthetic SceneFlow~\cite{mayer2016large} are used to distil a monocular model reaches higher performance than CRF-DGAN, but it requires both a large amount of additional stereo data for training and a more complex optimization process. 

\begin{figure}[!t]
\centering
\includegraphics[width=1\linewidth]{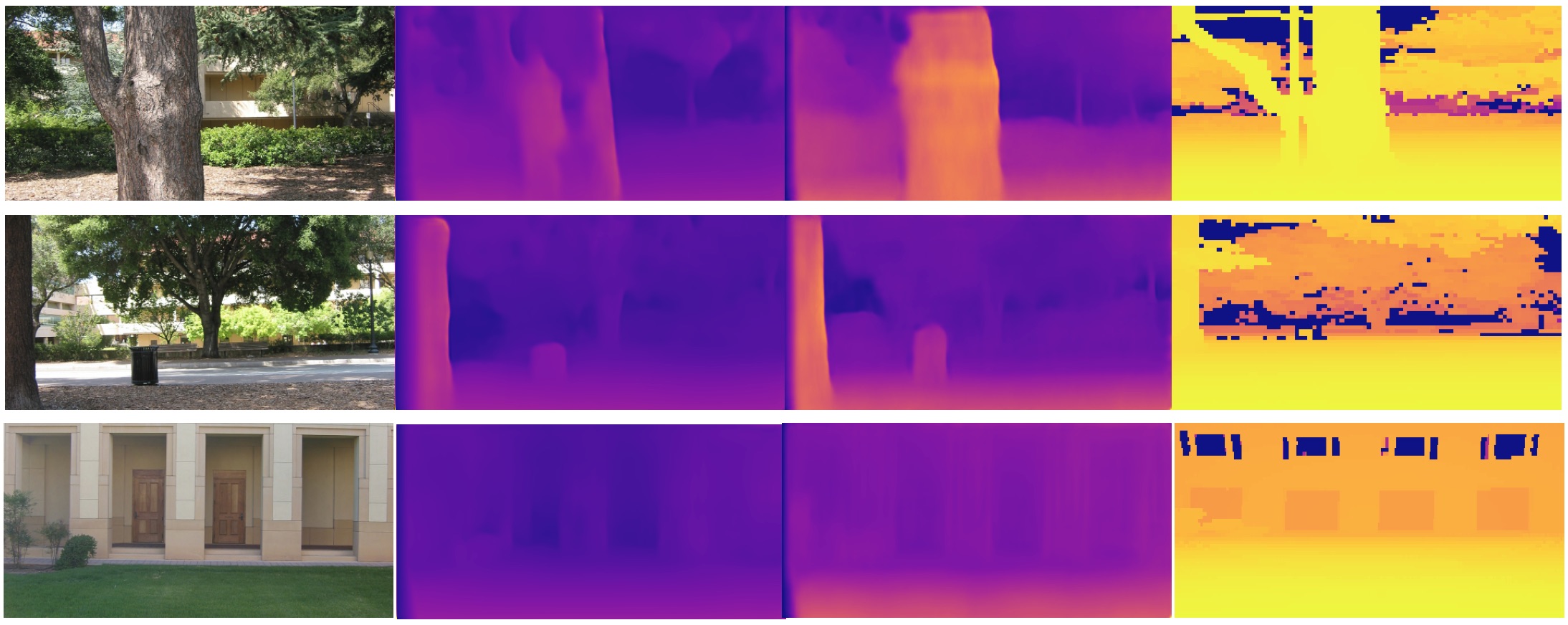} 
\caption{Examples of depth prediction results on the Make3D dataset. Qualitative comparison between structured and non-structured disparity maps is presented.}
\label{Make3d_comp}
\vspace{-5mm}
\end{figure}

\begin{figure}[!t]
\centering
\includegraphics[width=1\linewidth]{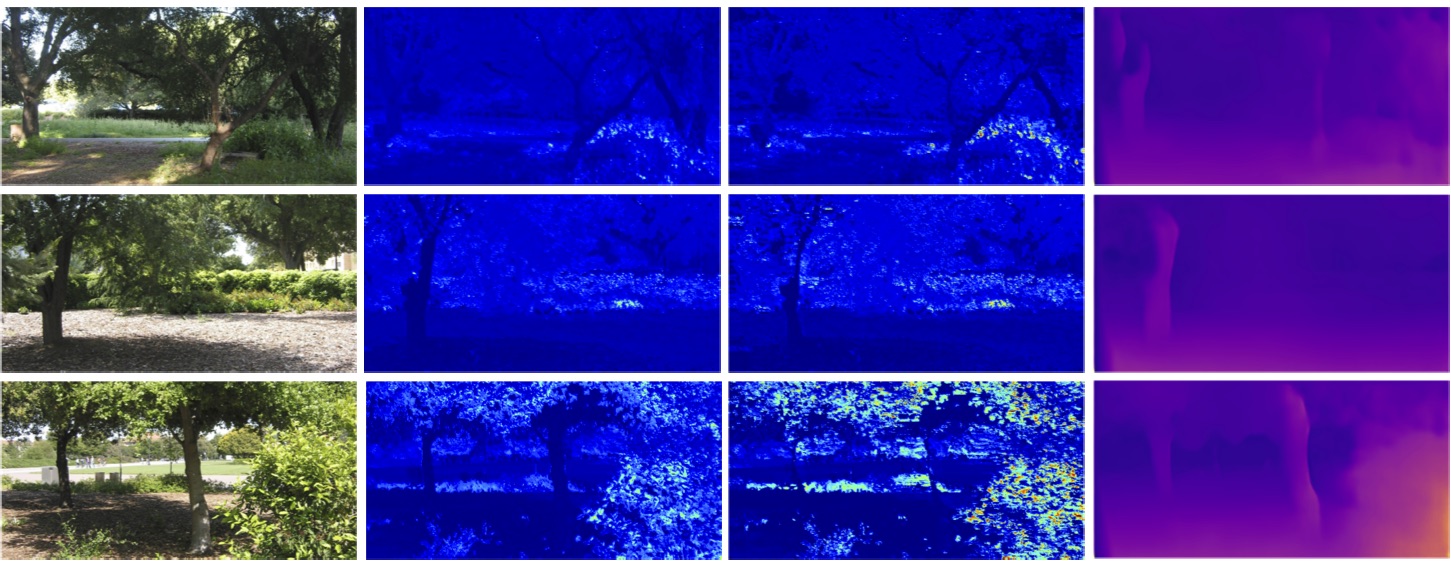} 
\caption{Examples of structured outputs of the real and the fake discriminative score-maps on the Make3D dataset, with the associated depth predictions.}
\label{Make3d_discr}
\vspace{-5mm}
\end{figure}

\noindent\textbf{Qualitative Analysis.} The performance can be qualitatively observed in Fig.~\ref{KITTI_comp} and~\ref{Cityscapes_comp} for KITTI and Cityscapes, respectively. The advantage of structured modeling between the generator and the discriminator can be observed in Fig.~\ref{KITTI_comp}, where our method is able to capture object details as well as objects in their entirety. Furthermore, we qualitatively evaluate a model learned on the Cityscapes dataset and tested on the Make3D dataset. The results are shown in Figure~\ref{Make3d_comp}. The importance of adding structural information when inferring on unfamiliar data can be clearly observed. Conditioning on the input images allows the approach to maintain a good detail consistency.
Figure~\ref{Make3d_discr} shows the structured output of the discriminative score-maps generated from associated real and synthesized samples. The areas in which the synthesized disparity values with low accuracy produce a high discriminative error.  Fig.~\ref{Cityscapes_comp} showcases different variants of the proposed CRF-DGAN approach and the improvement in quality. 
\vspace{0mm}




\noindent\textbf{Discussion on the Time Aspect.} On a single Titan V-100, with a batch size of 4, the model can infer 6 images with a resolution of $512 \times 256$ per second, which is near real-time speed. Further performance improvements in speed can be achieved through decreasing the size of the CRF receptive field and also consider approximation approaches in the expensive Gaussian convolutional operations, \eg~permutohedral lattice algorithm~\cite{adams2010fast}.

\section{Conclusion}
We have presented an end-to-end unsupervised deep learning framework for monocular depth estimation. The proposed framework consists of two generative adversarial sub-networks, aiming at on one hand generating distinct while complementary disparity maps, through accepting images from different views as input, and on the other hand, improving the generation quality via exploiting the adversarial learning strategy. We couple the dual-GAN by a deep CRF model, which is able to perform structured refinement and fusion of the predicted disparity maps from the generators and the adversarial scoremaps from the discriminators. The deep CRF coupling also makes the discriminator and the generator explicitly constrain on each other, and thus facilitates the optimization of the whole network for better disparity generation. We conducted extensive experiments on the challenging KITTI, Cityscapes, and Make3D datasets, clearly demonstrating the effectiveness of the proposed approach.

{\small
\bibliographystyle{ieee}
\bibliography{egbib}

\begin{thebibliography}{10}\itemsep=-1pt

\bibitem{adams2010fast}
A.~Adams, J.~Baek, and M.~A. Davis.
\newblock Fast high-dimensional filtering using the permutohedral lattice.
\newblock In {\em Computer Graphics Forum}, 2010.

\bibitem{behl2017bounding}
A.~Behl, O.~H. Jafari, S.~K. Mustikovela, H.~A. Alhaija, C.~Rother, and
  A.~Geiger.
\newblock Bounding boxes, segmentations and object coordinates: How important
  is recognition for 3d scene flow estimation in autonomous driving scenarios?
\newblock In {\em CVPR}, 2017.

\bibitem{Cordts2016Cityscapes}
M.~Cordts, M.~Omran, S.~Ramos, T.~Rehfeld, M.~Enzweiler, R.~Benenson,
  U.~Franke, S.~Roth, and B.~Schiele.
\newblock The cityscapes dataset for semantic urban scene understanding.
\newblock In {\em CVPR}, 2016.

\bibitem{eigen2015predicting}
D.~Eigen and R.~Fergus.
\newblock Predicting depth, surface normals and semantic labels with a common
  multi-scale convolutional architecture.
\newblock In {\em ICCV}, 2015.

\bibitem{eigen2014depth}
D.~Eigen, C.~Puhrsch, and R.~Fergus.
\newblock Depth map prediction from a single image using a multi-scale deep
  network.
\newblock In {\em NIPS}, 2014.

\bibitem{gan2018monocular}
Y.~Gan, X.~Xu, W.~Sun, and L.~Lin.
\newblock Monocular depth estimation with affinity, vertical pooling, and label
  enhancement.
\newblock In {\em ECCV}, 2018.

\bibitem{garg2016unsupervised}
R.~Garg, G.~Carneiro, and I.~Reid.
\newblock Unsupervised cnn for single view depth estimation: Geometry to the
  rescue.
\newblock In {\em ECCV}, 2016.

\bibitem{Geiger2013IJRR}
A.~Geiger, P.~Lenz, C.~Stiller, and R.~Urtasun.
\newblock Vision meets robotics: The kitti dataset.
\newblock {\em IJRR}, 2013.

\bibitem{godard2016unsupervised}
C.~Godard, O.~Mac~Aodha, and G.~J. Brostow.
\newblock Unsupervised monocular depth estimation with left-right consistency.
\newblock {\em CVPR}, 2017.

\bibitem{goodfellow2014generative}
I.~Goodfellow, J.~Pouget-Abadie, M.~Mirza, B.~Xu, D.~Warde-Farley, S.~Ozair,
  A.~Courville, and Y.~Bengio.
\newblock Generative adversarial nets.
\newblock In {\em NIPS}, 2014.

\bibitem{guo2018learning}
X.~Guo, H.~Li, S.~Yi, J.~Ren, and X.~Wang.
\newblock Learning monocular depth by distilling cross-domain stereo networks.
\newblock In {\em ECCV}, 2018.

\bibitem{koltun2011efficient}
P.~Kr{\"a}henb{\"u}hl and V.~Koltun.
\newblock Efficient inference in fully connected crfs with gaussian edge
  potentials.
\newblock {\em NIPS}, 2011.

\bibitem{adadepth}
J.~N. Kundu, P.~K. Uppala, A.~Pahuja, and R.~V. Babu.
\newblock Adadepth: Unsupervised content congruent adaptation for depth
  estimation.
\newblock In {\em CVPR}, 2018.

\bibitem{kuznietsov2017semi}
Y.~Kuznietsov, J.~St{\"u}ckler, and B.~Leibe.
\newblock Semi-supervised deep learning for monocular depth map prediction.
\newblock {\em CVPR}, 2017.

\bibitem{laina2016deeper}
I.~Laina, C.~Rupprecht, V.~Belagiannis, F.~Tombari, and N.~Navab.
\newblock Deeper depth prediction with fully convolutional residual networks.
\newblock {\em arXiv preprint arXiv:1606.00373}, 2016.

\bibitem{liu2016learningtpami}
F.~Liu, C.~Shen, G.~Lin, and I.~Reid.
\newblock Learning depth from single monocular images using deep convolutional
  neural fields.
\newblock {\em TPAMI}, 2016.

\bibitem{mahjourian2018unsupervised}
R.~Mahjourian, M.~Wicke, and A.~Angelova.
\newblock Unsupervised learning of depth and ego-motion from monocular video
  using 3d geometric constraints.
\newblock In {\em CVPR}, 2018.

\bibitem{mayer2016large}
N.~Mayer, E.~Ilg, P.~Hausser, P.~Fischer, D.~Cremers, A.~Dosovitskiy, and
  T.~Brox.
\newblock A large dataset to train convolutional networks for disparity,
  optical flow, and scene flow estimation.
\newblock In {\em CVPR}, 2016.

\bibitem{pilzer2019cvpr}
A.~Pilzer, S.~Lathuiliere, N.~Sebe, and E.~Ricci.
\newblock Refine and distill: Exploiting cycle-inconsistency and knowledge
  distillation for unsupervised monocular depth estimation.
\newblock In {\em CVPR}, 2019.

\bibitem{pilzer2018unsupervised}
A.~Pilzer, D.~Xu, M.~Puscas, E.~Ricci, and N.~Sebe.
\newblock Unsupervised adversarial depth estimation using cycled generative
  networks.
\newblock In {\em 3DV}, 2018.

\bibitem{ristovski2013continuous}
K.~Ristovski, V.~Radosavljevic, S.~Vucetic, and Z.~Obradovic.
\newblock Continuous conditional random fields for efficient regression in
  large fully connected graphs.
\newblock In {\em AAAI}, 2013.

\bibitem{saxena2006learning}
A.~Saxena, S.~H. Chung, and A.~Y. Ng.
\newblock Learning depth from single monocular images.
\newblock In {\em NIPS}, 2006.

\bibitem{saxena2009make3d}
A.~Saxena, M.~Sun, and A.~Y. Ng.
\newblock Make3d: Learning 3d scene structure from a single still image.
\newblock {\em TPAMI}, 2009.

\bibitem{turan2017non}
M.~Turan, Y.~Almalioglu, H.~Araujo, E.~Konukoglu, and M.~Sitti.
\newblock A non-rigid map fusion-based rgb-depth slam method for endoscopic
  capsule robots.
\newblock {\em arXiv preprint arXiv:1705.05444}, 2017.

\bibitem{wang2017learning}
C.~Wang, J.~M. Buenaposada, R.~Zhu, and S.~Lucey.
\newblock Learning depth from monocular videos using direct methods.
\newblock In {\em CVPR}, 2018.

\bibitem{wang2015towards}
P.~Wang, X.~Shen, Z.~Lin, S.~Cohen, B.~Price, and A.~Yuille.
\newblock Towards unified depth and semantic prediction from a single image.
\newblock In {\em CVPR}, 2015.

\bibitem{xie2015holistically}
S.~Xie and Z.~Tu.
\newblock Holistically-nested edge detection.
\newblock In {\em ICCV}, 2015.

\bibitem{xu2017multi}
D.~Xu, E.~Ricci, W.~Ouyang, X.~Wang, and N.~Sebe.
\newblock Multi-scale continuous {CRFs} as sequential deep networks for
  monocular depth estimation.
\newblock {\em CVPR}, 2017.

\bibitem{xu2018monocular}
D.~Xu, E.~Ricci, W.~Ouyang, X.~Wang, and N.~Sebe.
\newblock Monocular depth estimation using multi-scale continuous {CRFs} as
  sequential deep networks.
\newblock {\em TPAMI}, 41(6):1426--1440, 2018.

\bibitem{xu2018structured}
D.~Xu, W.~Wang, H.~Tang, H.~Liu, N.~Sebe, and E.~Ricci.
\newblock Structured attention guided convolutional neural fields for monocular
  depth estimation.
\newblock In {\em CVPR}, 2018.

\bibitem{zhan2018unsupervised}
H.~Zhan, R.~Garg, C.~S. Weerasekera, K.~Li, H.~Agarwal, and I.~Reid.
\newblock Unsupervised learning of monocular depth estimation and visual
  odometry with deep feature reconstruction.
\newblock {\em arXiv preprint arXiv:1803.03893}, 2018.

\bibitem{zhou2017unsupervised}
T.~Zhou, M.~Brown, N.~Snavely, and D.~G. Lowe.
\newblock Unsupervised learning of depth and ego-motion from video.
\newblock {\em CVPR}, 2017.

\bibitem{zou2018df}
Y.~Zou, Z.~Luo, and J.-B. Huang.
\newblock Df-net: Unsupervised joint learning of depth and flow using
  cross-task consistency.
\newblock In {\em ECCV}, 2018.

\end{thebibliography}
}

\end{document}